\documentclass[journal]{IEEEtran}
\IEEEoverridecommandlockouts

\usepackage[nocompress]{cite}
\usepackage{amsmath,amssymb,amsfonts}
\usepackage{algorithm}
\usepackage{algpseudocode}
\algrenewcommand\algorithmicreturn{\textbf{Return}}
\makeatletter
\newcommand{\algdash}[1]{\State\parbox[t]{\dimexpr\linewidth-\ALG@thistlm\relax}{\raggedright$-$\enspace#1\strut}}
\makeatother
\usepackage{graphicx}
\usepackage{textcomp}
\usepackage{xcolor}
\usepackage{booktabs}
\usepackage{multirow}
\usepackage{threeparttable} %
\usepackage{pifont} %
\usepackage{tikz}
\usetikzlibrary{arrows.meta, positioning, calc, fit, backgrounds}
\usepackage[scaled=0.92]{helvet} %
\usepackage{capt-of} %
\usepackage[hidelinks]{hyperref}
\usepackage{flushend} %

\makeatletter

\newcommand{\Rmnum}[1]{\expandafter\@slowromancap\romannumeral #1@}
\makeatother

\newcommand{\sorb}[1]{{\textcolor[rgb]{0.72,0.00,0.00}{\textbf{#1}}}}
\newcommand{\cmark}{\sorb{\ding{51}}}%
\newcommand{\xmark}{\ding{55}}%
\newcommand{\nrmark}{\textendash}%
\newlength\savedwidth
\newcommand{\whline}[1]{\noalign{\global\savedwidth\arrayrulewidth \global\arrayrulewidth #1}%
                   \hline \noalign{\global\arrayrulewidth\savedwidth}}

\graphicspath{{figures/}{figures/_intermediate/}}

\def\BibTeX{{\rm B\kern-.05em{\sc i\kern-.025em b}\kern-.08em
    T\kern-.1667em\lower.7ex\hbox{E}\kern-.125emX}}

\definecolor{revcolor}{rgb}{0,0,0}
\DeclareRobustCommand{\revtext}[1]{\leavevmode{\color{revcolor}#1}}
\begin{document}
\bstctlcite{IEEEexample:BSTcontrol}
\pagestyle{empty}
\title{LAC: Linear and Angular Compliance\\
for Humanoid Whole-body Control}

\author{Yang Liu, Zhongkai Gu, Wei Zhu, Mitsuhiro Hayashibe%
\thanks{All authors are with the Neuro-Robotics Laboratory, Tohoku University, Japan (e-mail: liu.yang.t3@dc.tohoku.ac.jp).}%
}

\maketitle
\thispagestyle{empty}

\begin{abstract}
\revtext{Physical interaction with objects and humans exposes humanoids to external forces and torques at various points on the body.}
\revtext{However, existing controllers either reject these loads or comply only with forces at a few links and ignore angular compliance.}
We present LAC, a general whole-body controller that simultaneously realizes commanded \underline{L}inear and \underline{A}ngular \underline{C}ompliance for wrenches applied to the upper body.
First, we synthesize whole-body compliant responses into a large-scale augmented dataset.
Sampled force and couple events are imposed on contact frames extracted from human interaction data. At each contact link, the external force and a virtual torque from the passively yielding kinematic chain drive a virtual admittance under the commanded stiffness.
Subsequently, teacher--student reinforcement learning trains a single policy to track the compliant motions under external wrenches.
Finally, simulation and real-world experiments demonstrate whole-body compliant responses to wrenches across the upper body, monotonic modulation over the full range of both stiffness commands, and applicability to teleoperated loco-manipulation tasks.
Code and demos are available at \url{https://lac-humanoid.github.io/}.
\end{abstract}

\section{Introduction}

\revtext{The focus of humanoid robotics research is shifting from imitating human motions to performing real-world tasks.}
Current motion-tracking controllers can accurately reproduce human movements on physical hardware~\cite{cheng2024expressive,liao2025beyondmimic}, and whole-body teleoperation systems extend these capabilities to interactive operation~\cite{he2024omnih2o,fu2024humanplus,ze2025twist}.
Real-world tasks such as loco-manipulation, however, go beyond tracking human-like motions and demand continuous force interaction with objects and people~\cite{fu2023deep,ben2025homie}.
Most existing controllers reject external forces as disturbances~\cite{pratt2006capture,radosavovic2024real,zhang2025falcon}.
FALCON~\cite{zhang2025falcon}, for instance, applies a torque-aware force curriculum so that the robot maintains tracking and withstands strong pulls.
Holding the reference is desirable when the task is to resist the force, but in tasks that require compliance, such as physical collaboration with a person or handling fragile objects, the same controller responds rigidly, producing uncontrolled forces.

By contrast, another line of work makes the robot comply with external wrenches, from classical impedance control to learning-based controllers on quadrupedal platforms and humanoids, as reviewed in Sec.~\ref{sec:related}.
Beyond the linear channel, real-world interaction often requires control over the attitude of the contact link.
In grasp-type interactions, such as a person twisting the robot's hand or the robot turning a door handle, the hand should hold its position while its attitude complies with the applied torque, calling for a low angular stiffness with a high linear one.
Carrying an object on the arms instead requires a high angular stiffness to keep the arms level.
Supporting both cases therefore requires independent control of linear and angular stiffness.

Realizing both channels simultaneously within a single learned policy requires large-scale training data containing compliant responses that are physically feasible and observable from proprioception.
For the linear channel, prior work~\cite{margolis2025softmimic} sets the displacement of the contact link proportional to the applied force, as if the link were a free rigid body, and this has proven effective.
For the angular channel, however, the rotation of a link under a wrench is the accumulated deflection of the joints behind it, and this deflection is also the main signal through which the policy perceives the wrench.
Suppose the target rotation were set proportional to the torque of the force about the center of mass (CoM) or another fixed point of the link.
Then two identical forces applied on opposite sides of that point would produce opposite torques and hence opposite target rotations, while deflecting the joints similarly.
\revtext{Such a target is not guaranteed to be physically feasible or observable; indeed, a policy trained on such targets reproduces them with large errors.}
Consequently, no existing policy realizes commanded linear and angular compliance for wrenches across the upper body.

We present LAC to close this gap.
To synthesize training data, we first extract contact frames from human--object and human--human interaction data.
For each contact frame, a sampled force or couple event is applied, and a virtual admittance under the commanded stiffness, realized through whole-body inverse kinematics, generates a feasible compliant motion trajectory for the entire body.
In particular, driving the angular channel through the passively yielding kinematic chain anchors the angular stiffness command to a physical reference.
Finally, teacher--student reinforcement learning trains a single policy on the augmented data.
We validate the framework extensively in simulation and on a real 23-DoF Unitree G1 humanoid robot.

In summary, LAC makes the following contributions:
\begin{itemize}
\item A general learning-based controller that responds compliantly with whole-body degrees of freedom to one or more wrenches acting at different locations of the upper body;
\item To the best of our knowledge, the first method that effectively achieves commanded linear and angular compliance on a humanoid robot, verified in simulation and on the real robot.
\end{itemize}

\section{Related Work}
\label{sec:related}

Compliance control has several classical methods.
Hybrid position/force control regulates the contact force directly~\cite{raibert1981hybrid}, while impedance control prescribes the resulting deviation through a commanded stiffness~\cite{hogan1984impedance}.
Letting a learned policy modulate the impedance parameters has further proven effective for contact-rich manipulation~\cite{martin2019variable}.
Reinforcement learning has recently brought compliance to legged platforms.
Quadrupedal policies learn compliant responses by relaxing the tracking rewards under perturbation~\cite{hartmann2024deep}, learn explicit end-effector force control~\cite{portela2024learning,zhi2025learning}, or track the reference of a commanded virtual impedance~\cite{xu2025facet}.

\begin{table}[t]
	\centering
    \fontsize{6pt}{8pt}\selectfont
	\renewcommand{\arraystretch}{1.15}
  \setlength\tabcolsep{0.38mm}
  \caption{Capability comparison with prior work.
  }
	\label{tab:capability}
  \begin{threeparttable}
	\begin{tabular}{l|c|c|c|c|c} \whline{1pt}
     \multirow{2}{*}{Method} & \multirow{2}{*}{Wrench Application Site} & Whole-Body & Linear & Angular & Single \\
	   & & Response & Compliance & Compliance & Policy \\ \whline{0.7pt}
       SoftMimic~\cite{margolis2025softmimic} & Hands & \cmark & \cmark & \nrmark & \xmark \\
       GentleHumanoid~\cite{lu2025gentlehumanoid} & Both arms & \xmark & \cmark & \xmark & \cmark \\
        CHIP~\cite{chen2025chip} & Hands & \cmark & \cmark & \xmark & \cmark \\
       \textbf{LAC}~(\textbf{Ours}) & Upper body & \cmark & \cmark & \cmark & \cmark \\
	  \whline{1pt}
	\end{tabular}
  \begin{tablenotes}
  \item \!\!\!\!\!\!``\nrmark'' indicates that no supporting experimental evidence is reported in the
  original work.
  \end{tablenotes}
  \end{threeparttable}
\end{table}

Several recent efforts bring compliance to humanoids, summarized in Table~\ref{tab:capability}.
SoftMimic~\cite{margolis2025softmimic} introduces compliant motion augmentation, in which whole-body inverse kinematics synthesizes offline how a reference motion should deform under an external wrench and a commanded stiffness, and the policy is rewarded for reproducing the synthesized response.
Each policy is trained for one motion clip with the wrench applied to one hand link at a time, and the angular stiffness command is not yet supported by experimental evidence.
GentleHumanoid~\cite{lu2025gentlehumanoid} integrates impedance control into a general motion-tracking policy through a unified spring model over the arm links, keeping interaction forces within a tunable safety threshold.
The compliance is linear, and the lower body maintains balance rather than joining the compliant response.
CHIP~\cite{chen2025chip} applies forces at the end-effectors during motion tracking training, computes hindsight tracking goals for the policy observation, and trains with a reward that tracks the original goals.
The compliance is limited to the linear response of the end-effectors.
In summary, existing humanoid controllers confine compliance to a few body links, and commanded angular compliance remains to be achieved.

\section{Method}

\begin{figure*}[t]
\centering
\includegraphics[width=0.95\textwidth]{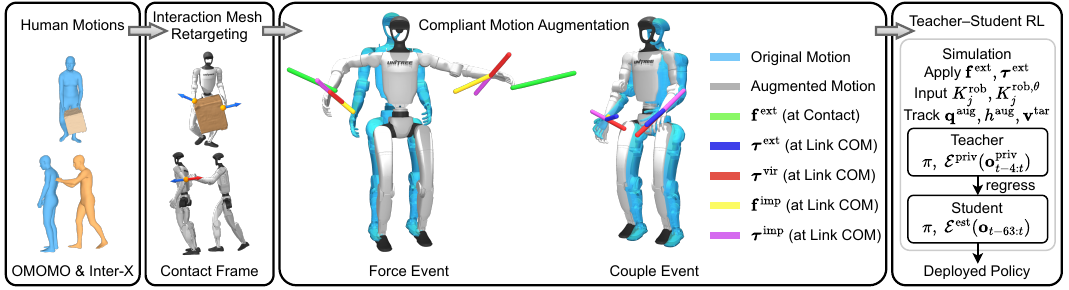}
\caption{\revtext{\textbf{Overview of LAC.}}
\revtext{Human interaction data are retargeted to the humanoid to obtain contact frames.}
\revtext{Force or couple events are imposed on these frames.} \revtext{A virtual admittance under the commanded stiffness, realized through whole-body inverse kinematics, generates the compliant response (Sec.~\ref{ssec:aug}).}
\revtext{Teacher--student reinforcement learning in simulation then trains a single policy on the augmented samples (Sec.~\ref{ssec:rl}).}}
\label{fig:ik}
\end{figure*}

\subsection{Problem Definition}
In this paper, we consider the problem of humanoid loco-manipulation with adjustable linear and angular stiffness.
The robot receives a velocity command $\mathbf{v}^{\mathrm{cmd}}=({v}^\mathrm{cmd}_{x}, {v}^\mathrm{cmd}_{y},{\omega}^\mathrm{cmd}_{z})$, a base height command $h^{\mathrm{cmd}}$,
and upper-body reference joint angles $\mathbf{q}^{\mathrm{ref}} \in \mathbb{R}^{11}$.
In addition, the robot is given a five-dimensional stiffness command that specifies how compliantly it should react to external wrenches:
a linear--angular stiffness pair $(K_{\mathrm{L}}^{\mathrm{rob}},K_{\mathrm{L}}^{\mathrm{rob},\theta})$ shared by all left-arm links, a pair $(K_{\mathrm{R}}^{\mathrm{rob}},K_{\mathrm{R}}^{\mathrm{rob},\theta})$ shared by all right-arm links, and a linear stiffness $K_{\mathrm{T}}^{\mathrm{rob}}$ for the torso, which has no angular channel.
We write $K_j^{\mathrm{rob}}$ and $K_j^{\mathrm{rob},\theta}$ for the values assigned to upper-body link $j$ through its group.
During operation, one or more external wrenches $(\mathbf{f}_j^{\mathrm{ext}},\boldsymbol{\tau}_j^{\mathrm{ext}})$, which the robot cannot directly observe, act on its upper-body links.
\revtext{The linear stiffness modulates the CoM displacement of the contact link, and the angular stiffness its attitude change.}
The robot outputs whole-body joint angles $\mathbf{q}\in\mathbb{R}^{23}$ that maintain balanced locomotion, while the base velocity, base height, and upper-body posture deviate from the commands as dictated by the wrenches and the commanded stiffness.

\subsection{Compliant Motion Augmentation}
\label{ssec:aug}%

\textbf{Data preprocessing.}
Our augmentation starts from a set of contact frames $\mathcal{L}$ distilled from human interaction data, as illustrated by the first two blocks of Fig.~\ref{fig:ik}.
We retarget the human--object interactions of OMOMO~\cite{li2023object} and the human--human interactions of Inter-X~\cite{xu2024inter} to the humanoid with OmniRetarget~\cite{yang2025omniretarget}.
\revtext{For Inter-X, we retarget the two actors separately and then retarget one actor again using the other robot as the object, which suppresses their interpenetration.}
From the retargeted clips we keep exactly the frames in which at least one of the eleven upper-body links touches the object or the partner robot.
Each of these frames is stored in $\mathcal{L}$ as a contact frame
$e=(\mathbf{q}^{\mathrm{ref}},m_j,\mathbf{p}_j,\hat{\mathbf{n}}_j)$:
the upper-body joint angles $\mathbf{q}^{\mathrm{ref}}\in\mathbb{R}^{11}$, which serve as the reference command during augmentation, a contact activation mask $m_j\in\{0,1\}$, and, for each contacted link, the contact point $\mathbf{p}_j\in\mathbb{R}^3$ and the inward surface normal $\hat{\mathbf{n}}_j$, both expressed in the link frame.
The augmentation described below samples contact frames $e\sim\mathcal{L}$.

\textbf{Compliant response synthesis.}
Given a sampled contact frame $e\sim\mathcal{L}$, Algorithm~\ref{alg:cma} synthesizes one augmented episode.
We first build a $T$-frame reference trajectory representing the motion the robot would perform under no external wrench.
The base is held at the position $\mathbf{p}_b^{\mathrm{ref}}=(0,0,h^{\mathrm{cmd}})^\top$ with identity orientation $R_b^{\mathrm{ref}}=I_3$, where the height command is sampled as $h^{\mathrm{cmd}}\sim\mathcal{U}(h_{\min},h_{\max})$.
The legs stay at the default stance, and the upper body is frozen at the contact frame's reference joint angles $\mathbf{q}^{\mathrm{ref}}$.
The reference pose of contact link $j$ is then obtained by forward kinematics,
\begin{equation}
(\mathbf{x}_j^{\mathrm{ref}},R_j^{\mathrm{ref}})=\mathrm{FK}_j(\mathbf{p}_b^{\mathrm{ref}},R_b^{\mathrm{ref}},\mathbf{q}^{\mathrm{ref}}),
\end{equation}
\revtext{where $\mathbf{x}_j^{\mathrm{ref}}$ and $R_j^{\mathrm{ref}}$ are the CoM position and orientation of the link in the world frame.}
Locomotion commands are zero during synthesis; walking is reintroduced when the augmented data is used in policy training (Sec.~\ref{ssec:rl}).

Onto this reference we impose wrench events, one at each active contact point.
Within an episode all events are of the same type, force events or couple events.
The third block of Fig.~\ref{fig:ik} shows one of each type together with the augmented motion it produces.
The events share one trapezoidal time profile $\Pi(t)\in[0,1]$, which rises from zero at the event-start frame $t_s$, holds at one, and releases (phase durations resampled per episode; Table~\ref{tab:aug_params}).
Each event has its own peak wrench.
A force event ($\boldsymbol{\tau}_j^{\mathrm{ext}}\equiv \mathbf{0}$) pushes along the direction $\hat{\mathbf{d}}_j=R_j^{\mathrm{ref}}\,\hat{\mathbf{n}}_j$, the stored contact normal rotated to the world frame by the reference orientation and held fixed for the whole event.
Instead of sampling the force magnitude directly, we follow SoftMimic~\cite{margolis2025softmimic} and displace the setpoint of a virtual environment spring of sampled stiffness $K_j^{\mathrm{env}}$ by $\Delta x_j$, sampled from the $\mathrm{Beta}(3,1)$ distribution on $[x_j^{\min},x_j^{\max}]$, where $x_j^{\max}=\min\big(F_{\lim}/K_j^{\mathrm{env}},\ d_{\lim}K_j^{\mathrm{rob}}/K_j^{\mathrm{env}}\big)$:
\begin{equation}
\mathbf{F}_j^{\mathrm{peak}}=K_j^{\mathrm{env}}\,\Delta x_j\,\hat{\mathbf{d}}_j,\qquad
\mathbf{f}_j^{\mathrm{ext}}(t)=m_j\,\Pi(t)\,\mathbf{F}_j^{\mathrm{peak}}.
\end{equation}
The bound $x_j^{\max}$ caps the peak force $\|\mathbf{F}_j^{\mathrm{peak}}\|$ at $F_{\lim}$ and the commanded quasi-static deviation $\|\mathbf{F}_j^{\mathrm{peak}}\|/K_j^{\mathrm{rob}}$ at $d_{\lim}$ by construction.
A couple event ($\mathbf{f}_j^{\mathrm{ext}}\equiv \mathbf{0}$) complements the force events, covering grasp-type contact, which applies a couple independent of any application point.
The applied couple is sampled as
\begin{equation}
\boldsymbol{\tau}_j^{\mathrm{ext}}(t)=m_j\,\Pi(t)\,\tau_j^{\mathrm{peak}}\,\hat{\mathbf{c}}_j,
\end{equation}
where the peak magnitude is $\tau_j^{\mathrm{peak}}=\tau_{\mathrm{s}}\,\beta_j$ with $\beta_j\sim\mathrm{Beta}(3,1)$, and the direction $\hat{\mathbf{c}}_j$ is sampled either uniformly on the unit sphere or within a narrow cone about the contact link's own joint axis at the event start, with the sign of the axis chosen at random (Table~\ref{tab:aug_params}).

The applied wrench does not drive the angular response directly.
The contact link is not a free rigid body but part of the robot's kinematic chain, so how far it can rotate under a given wrench is dictated by the joints behind it.
We therefore compute the rotation that the wrench produces through the passively yielding chain.
The force is mapped to upper-body joint torques through the translational Jacobian $J_{p,j}$ of its application point, and the couple is mapped through the rotational Jacobian $J_{r,j}$ of the contact link.
\revtext{Assuming fixed joint targets, the joints deflect quasi-statically according to the real robot's joint-level PD stiffness $\Lambda_q$:}
\begin{equation}
\boldsymbol{\tau}_u=\sum_j\Big(J_{p,j}^{\top}\,\mathbf{f}_j^{\mathrm{ext}}+J_{r,j}^{\top}\,\boldsymbol{\tau}_j^{\mathrm{ext}}\Big),\qquad
\Delta\mathbf{q}_u=\Lambda_q^{-1}\,\boldsymbol{\tau}_u.
\end{equation}
Forward kinematics at the deflected posture then gives each contact link's orientation under the wrench:
\begin{equation}
(\,\cdot\,,R_j^{\Delta})=\mathrm{FK}_j\big(\mathbf{p}_b^{\mathrm{ref}},\,R_b^{\mathrm{ref}},\,\mathbf{q}^{\mathrm{ref}}+\Delta\mathbf{q}_u\big).
\end{equation}
Its deviation from the reference orientation, $\Delta\boldsymbol{\theta}_j=\mathrm{Log}\!\big(R_j^{\Delta}\,(R_j^{\mathrm{ref}})^{\top}\big)$, is the passive rotation of the link under the wrench and serves as the reference for the angular response.
We convert it into the virtual torque driving the angular admittance,
\begin{equation}
\boldsymbol{\tau}_j^{\mathrm{vir}}=\kappa\,\Delta\boldsymbol{\theta}_j.
\end{equation}
The Jacobians and application points are evaluated at the previous frame's solved configuration, while the deflection $\Delta\mathbf{q}_u$ is applied from the reference posture.
The chain behind the torso contains only the single waist-yaw joint, so the torso can barely rotate under a wrench.
We therefore do not design an angular stiffness command for it, and torso contacts use the linear channel only.

\begin{algorithm}[t]
\caption{Compliant Response Synthesis}
\label{alg:cma}
\footnotesize
\begin{algorithmic}
\Require Contact frame $e=(\mathbf{q}^{\mathrm{ref}},m_j,\mathbf{p}_j,\hat{\mathbf{n}}_j)\sim\mathcal{L}$
\Ensure Augmented trajectory: per-frame $\mathbf{q}^{\mathrm{aug}}(t)$, $\mathbf{v}^{\mathrm{aug}}(t)$, $h^{\mathrm{aug}}(t)$, $\mathbf{f}_j^{\mathrm{ext}}(t)$, $\boldsymbol{\tau}_j^{\mathrm{ext}}(t)$ for $t=1,\dots,T$, and robot stiffness $K_j^{\mathrm{rob}}$, $K_j^{\mathrm{rob},\theta}$; or $\varnothing$ if every attempt fails
\For{$\mathrm{iter}=1$ \textbf{to} $N_{\mathrm{iter}}$}
  \algdash{Initialize IK state with $h^{\mathrm{cmd}}\sim\mathcal{U}(h_{\min},h_{\max})$ and $\mathbf{q}^{\mathrm{ref}}$}
  \algdash{Sample trapezoidal time profile $\Pi(t)$, environment stiffness $K_j^{\mathrm{env}}$, robot stiffness $K_j^{\mathrm{rob}},K_j^{\mathrm{rob},\theta}$, and the wrench events (force-field displacement $\Delta x_j$ for force events; couple magnitude $\tau_j^{\mathrm{peak}}$ and direction $\hat{\mathbf{c}}_j$ for couple events)}
  \For{$\mathrm{decay}=1$ \textbf{to} $N_{\mathrm{decay}}$}
    \For{$t=1$ \textbf{to} $T$}
      \algdash{Compute external wrench $\mathbf{f}_j^{\mathrm{ext}}(t),\boldsymbol{\tau}_j^{\mathrm{ext}}(t)$ and virtual torque $\boldsymbol{\tau}_j^{\mathrm{vir}}(t)$}
      \algdash{Compute impedance driving force $\mathbf{f}_j^{\mathrm{imp}}(t)$ and torque $\boldsymbol{\tau}_j^{\mathrm{imp}}(t)$}
      \algdash{Compute target pose $(\mathbf{x}_j^{\mathrm{vir}}(t),R_j^{\mathrm{vir}}(t))$ of each contact link}
      \algdash{Solve IK with updated target pose}
    \EndFor
    \If{the peak wrench is below the weak floor}
      \algdash{\textbf{Break}}
    \ElsIf{the trajectory is not feasible}
      \algdash{Reduce the external wrench $(\mathbf{f}_j^{\mathrm{ext}},\boldsymbol{\tau}_j^{\mathrm{ext}})\gets\rho\,(\mathbf{f}_j^{\mathrm{ext}},\boldsymbol{\tau}_j^{\mathrm{ext}})$}
    \Else
      \algdash{\algorithmicreturn{}~trajectory and robot stiffness}
    \EndIf
  \EndFor
\EndFor
\algdash{\algorithmicreturn{}~$\varnothing$}
\end{algorithmic}
\end{algorithm}

\revtext{Each active contact link then has a virtual second-order system at its CoM, with mass $M$ and isotropic inertia $I$, that integrates the driving signals into a target pose in the world frame.}
With $(\mathbf{x}_j,R_j)$ the link's currently solved CoM pose, $(\dot{\mathbf{x}}_j,\boldsymbol{\omega}_j)$ its linear and angular velocity, and its deviations from the reference $\mathbf{e}_j^{x}=\mathbf{x}_j^{\mathrm{ref}}-\mathbf{x}_j$ and $\mathbf{e}_j^{R}=\mathrm{Log}(R_j^{\mathrm{ref}}R_j^{\top})$, an impedance wrench pulls the link back toward its reference pose with exactly the commanded stiffness:
\begin{equation}
\mathbf{f}_j^{\mathrm{imp}}=K_j^{\mathrm{rob}}\mathbf{e}_j^{x}-K_d\,\dot{\mathbf{x}}_j,\quad \boldsymbol{\tau}_j^{\mathrm{imp}}=K_j^{\mathrm{rob},\theta}\mathbf{e}_j^{R}-K_d^{\theta}\,\boldsymbol{\omega}_j,
\end{equation}
where $K_d=2\zeta\sqrt{M K_j^{\mathrm{rob}}}$ and $K_d^{\theta}=2\zeta\sqrt{I K_j^{\mathrm{rob},\theta}}$ set critical damping ($\zeta=1$).
The virtual dynamics balances this restoring wrench against the external drive, the force $\mathbf{f}_j^{\mathrm{ext}}$ on the linear channel and the virtual torque $\boldsymbol{\tau}_j^{\mathrm{vir}}$ on the angular channel, with $D$ and $D_\omega$ damping the virtual state for numerical stability:
\begin{align}
M\ddot{\mathbf{x}}_j^{\mathrm{vir}}&=\mathbf{f}_j^{\mathrm{imp}}+\mathbf{f}_j^{\mathrm{ext}}-D\dot{\mathbf{x}}_j^{\mathrm{vir}},\\
I\dot{\boldsymbol{\omega}}_j^{\mathrm{vir}}&=\boldsymbol{\tau}_j^{\mathrm{imp}}+\boldsymbol{\tau}_j^{\mathrm{vir}}-D_\omega\boldsymbol{\omega}_j^{\mathrm{vir}}.
\end{align}
At steady state the impedance torque balances the virtual torque at $\mathbf{e}_j^{R}=-(\kappa/K_j^{\mathrm{rob},\theta})\,\Delta\boldsymbol{\theta}_j$.
\revtext{Commanding $K_j^{\mathrm{rob},\theta}=\kappa$ thus sets the target rotation to the passive rotation, smaller values command a larger rotation, and larger values command a smaller one.}
\revtext{The command $K_j^{\mathrm{rob},\theta}$ is therefore a stiffness relative to the passive chain, not the ratio of the external torque to the rotation.}
The virtual velocities $\dot{\mathbf{x}}_j^{\mathrm{vir}},\boldsymbol{\omega}_j^{\mathrm{vir}}$ are integrated by one explicit Euler step of size $\Delta t$ and advance the target pose from the currently solved pose,
\begin{equation}
\mathbf{x}_j^{\mathrm{vir}}=\mathbf{x}_j+\dot{\mathbf{x}}_j^{\mathrm{vir}}\Delta t,\qquad R_j^{\mathrm{vir}}=\mathrm{Exp}(\boldsymbol{\omega}_j^{\mathrm{vir}}\Delta t)\,R_j,
\end{equation}
which becomes the tracking target of the inverse kinematics below.
An active torso contact keeps its orientation target at the reference pose.

A whole-body differential inverse kinematics step realizes the targets on the robot.
Writing $\bar{\mathbf{q}}=(\mathbf{p}_b,R_b,\mathbf{q})$ for the full configuration, with base position $\mathbf{p}_b\in\mathbb{R}^3$, base orientation $R_b\in\mathrm{SO}(3)$, and joint angles $\mathbf{q}=(\mathbf{q}_u,\mathbf{q}_{\mathrm{leg}})\in\mathbb{R}^{23}$, split into upper-body and leg components, each frame solves
\begin{align}
\bar{\mathbf{q}}^{*}={}&\arg\min_{\bar{\mathbf{q}}}\ {\textstyle\sum_j} m_j\Pi(t)\big[w_1^{p}\|\mathbf{x}_j^{\mathrm{vir}}-\mathbf{x}_j(\bar{\mathbf{q}})\|^2+{}\notag\\
&\hspace{-1.2em}w_1^{o}\|\mathrm{Log}(R_j^{\mathrm{vir}}R_j(\bar{\mathbf{q}})^{\top})\|^2\big]+w_{b}^{p}\|\mathbf{p}_{b}^{\mathrm{ref}}-\mathbf{p}_{b}\|^2+{}\notag\\
&\hspace{-1.2em}w_{b}^{o}\|\mathrm{Log}(R_{b}^{\mathrm{ref}}R_{b}^{\top})\|^2+w_{\mathrm{post}}\|\mathbf{q}^{\mathrm{ref}}-\mathbf{q}_u\|^2\\
\mathrm{s.t.}\ \ &\mathbf{q}^{-}\le \mathbf{q}\le \mathbf{q}^{+},\;\; h_{\min}\le p_{b,z}\le h_{\max},\;\; \phi_b=\theta_b=0,\notag\\
&\hspace{-1.2em}\mathbf{q}_{\mathrm{leg}}=\mathbf{q}_{\mathrm{leg}}^{\mathrm{ref}},\;\; |\dot p_{b,x}|,\,|\dot p_{b,y}|\le\bar v_{xy},\;\; |\dot p_{b,z}|\le\bar v_z,\;\; |\dot\psi_b|\le\bar\omega_z\notag
\end{align}
where $\mathbf{x}_j(\bar{\mathbf{q}})$ and $R_j(\bar{\mathbf{q}})$ are the CoM position and orientation of contact link $j$ at the configuration $\bar{\mathbf{q}}$.
The first term tracks each contact link's admittance target, with position and orientation weights $w_1^{p}$ and $w_1^{o}$ gated by the mask $m_j$ and ramped by the time profile $\Pi(t)$.
The second and third terms anchor the base to its reference pose with weights $w_{b}^{p}$ and $w_{b}^{o}$, and the fourth term regularizes the upper-body joints $\mathbf{q}_u$ toward $\mathbf{q}^{\mathrm{ref}}$ with weight $w_{\mathrm{post}}$.
The weights are ordered $w_1\gg w_{b}>w_{\mathrm{post}}$ (Table~\ref{tab:aug_params}), so under a wrench the optimizer lets the joint angles deviate from $\mathbf{q}^{\mathrm{ref}}$ first and moves the base next, while the contact targets remain closely tracked.
The constraints keep the joint angles within their ranges $[\mathbf{q}^{-},\mathbf{q}^{+}]$ and the base height $p_{b,z}$ within the command range $[h_{\min},h_{\max}]$.
The leg joints are fixed at the stance values $\mathbf{q}_{\mathrm{leg}}^{\mathrm{ref}}$, and the base roll $\phi_b$ and pitch $\theta_b$ are fixed to zero.
The planar, vertical, and yaw rates of the base (with $\psi_b$ the base yaw) are capped at $\bar v_{xy}$, $\bar v_z$, and $\bar\omega_z$.
\revtext{In practice, each frame takes one step toward $\bar{\mathbf{q}}^{*}$.} \revtext{A quadratic program over the rate of $\bar{\mathbf{q}}$ minimizes the objective linearized at the current configuration, and the rate is integrated over $\Delta t$.}
\revtext{The augmented motion consists of components of $\bar{\mathbf{q}}$ and its rate.} \revtext{These are the upper-body joint angles $\mathbf{q}^{\mathrm{aug}}(t)=\mathbf{q}_u$, the base height $h^{\mathrm{aug}}(t)=p_{b,z}$, and the base velocity $\mathbf{v}^{\mathrm{aug}}(t)$ from $(\dot p_{b,x},\dot p_{b,y},\dot\psi_b)$.}
The base velocity and the external wrenches are computed in the world frame and rotated into the base frame for storage.

Finally, the episode must pass a feasibility check (Algorithm~\ref{alg:cma}).
The synthesized motion is kept only if it is consistent with the commanded compliance.
During the wrench event, the impedance driving wrench must balance the external wrench, requiring $\mathrm{mean}_t\big|\,\|\mathbf{f}_j^{\mathrm{imp}}\|-\|\mathbf{f}_j^{\mathrm{ext}}\|\,\big|\le 5$~N on the force side and $\mathrm{mean}_t\big|\,\|\boldsymbol{\tau}_j^{\mathrm{imp}}\|-\|\boldsymbol{\tau}_j^{\mathrm{vir}}\|\,\big|\le 1.5$~N\,m on the torque side.
A larger residual means the optimizer could not realize the commanded response, and the stored data would misrepresent the interaction.
The motion must also remain free of self-collision and keep the total applied wrench within budget (Table~\ref{tab:aug_params}).
A failed check does not discard the episode.
Instead, every active wrench is scaled down by one of the gate-specific decay factors $\rho_1$--$\rho_3$ and the synthesis is repeated, so a wrench too strong to be feasible is reduced until it matches the strongest response the robot can realize.
The attempt is abandoned only when the decayed wrench falls below a minimum threshold (Table~\ref{tab:aug_params}); a fresh restart then resamples the whole episode.
Each contact frame is allowed at most $N_{\mathrm{iter}}$ restarts, each with at most $N_{\mathrm{decay}}$ decay retries (Algorithm~\ref{alg:cma}).
Each accepted episode stores the augmented motion, the applied wrench profiles, and the stiffness commands; the collection over $\mathcal{L}$ forms the dataset used next.

\begin{table}[t]
\centering
\caption{Compliant motion augmentation parameters.}
\label{tab:aug_params}
\setlength{\tabcolsep}{2.5pt}\renewcommand{\arraystretch}{1.15}\scriptsize
\begin{tabular}{llll}
\hline
Symbol & Meaning & Value/Range & Units\\
\hline
$T,\Delta t$ & frames / time step & 500, 0.02 & --, s\\
$h_{\min},h_{\max}$ & base height range & 0.56, 0.78 & m\\
-- & rest; ramp, hold, release & $\mathcal{U}(0.5,1)$; $\mathcal{U}(1,3)$ & s\\
$K_j^{\mathrm{env}}$ & env.\ stiffness & log-$\mathcal{U}[10,500]$ & N/m\\
$x_j^{\min}$ & min.\ setpoint disp. & 0.01 & m\\
$F_{\lim},d_{\lim}$ & force / disp.\ caps & 70, 4.0 & N, m\\
$K_j^{\mathrm{rob}}$ & robot lin.\ stiffness & log-$\mathcal{U}[10,500]$ & N/m\\
$\tau_{\mathrm{s}}$ & couple sampling scale & 5 & N\,m\\
-- & axis-bias prob.\ / cone & 0.5, $10^\circ$ & --\\
$\Lambda_q$ & joint PD stiffness (waist/arm) & 40.2, 14.3 & N\,m/rad\\
$\kappa$ & virtual torque gain & 30 & N\,m/rad\\
$K_j^{\mathrm{rob},\theta}$ & robot ang.\ stiffness & log-$\mathcal{U}[10,100]$ & N\,m/rad\\
$M,I$ & virtual mass / inertia & 1.0, 1.0 & kg, kg\,m$^2$\\
$\zeta$ & damping ratio & 1.0 & --\\
$D$ & virtual damping (lin.) & 2.0 & N\,s/m\\
$D_\omega$ & virtual damping (ang.) & 2.0 & N\,m\,s/rad\\
$w_1^{p},w_1^{o}$ & contact task weights & 6, 6 & --\\
$w_{b}^{p},w_{b}^{o}$ & base anchor weights & 1.5, 0.75 & --\\
$w_{\mathrm{post}}$ & posture weight & 0.2 & --\\
$\mathbf{q}^{-},\mathbf{q}^{+}$ & joint-angle bounds & $0.95\times$ model range & rad\\
$\bar v_{xy},\bar v_z,\bar\omega_z$ & base vel.\ caps & 0.6, 0.5, 0.6 & m/s, rad/s\\
-- & sum caps (force / couple) & 70, 10 & N, N\,m\\
$\rho_1,\rho_2,\rho_3$ & decay factors & 0.8, 0.5, 0.8 & --\\
-- & weak floors (force / couple) & 15, 0.5 & N, N\,m\\
$N_{\mathrm{iter}},N_{\mathrm{decay}}$ & restarts / decay retries & 10, 50 & --\\
\hline
\end{tabular}
\end{table}

\textbf{Dataset composition.}
\revtext{We solve the inverse kinematics with mink~\cite{Zakka_Mink_Python_inverse_2026} on a MuJoCo~\cite{todorov2012mujoco} model of the robot.}
\revtext{The resulting dataset contains 378{,}051 ten-second episodes (about 1{,}050 hours), 278{,}258 with force events and 99{,}793 with couple events.}
Couple events act on arm links only.
\revtext{Algorithm~\ref{alg:cma} returns an episode for 97.5\% of the contact frames.}
\revtext{Counting restarts and decay retries, its loop over the $T$ frames runs about four times on average for each accepted episode.}
\revtext{The synthesis ran in up to 368 parallel single-threaded processes on Intel Xeon CPUs and took about 3{,}000 CPU-hours in total.}

\subsection{Reinforcement Learning with Augmented Samples}
\label{ssec:rl}

\textbf{Observations.}
\revtext{During policy training each environment replays augmented episodes, as illustrated by the fourth block of Fig.~\ref{fig:ik}.}
The stored wrench profiles are applied to the contact links of the simulated robot, and the episode supplies both the commands the robot receives and the motion it should realize.
The augmented motion is used in command--residual form, where each tracked quantity splits into the command the operator would send and the wrench-driven deviation from it.
The upper-body joint angles split into $\mathbf{q}^{\mathrm{ref}}$ and $\mathbf{q}^{\mathrm{res}}=\mathbf{q}^{\mathrm{aug}}-\mathbf{q}^{\mathrm{ref}}$, and the base height into $h^{\mathrm{cmd}}$ and $h^{\mathrm{res}}=h^{\mathrm{aug}}-h^{\mathrm{cmd}}$.
Their tracking targets are $\mathbf{q}^{\mathrm{aug}}$ and $h^{\mathrm{aug}}$.
The base velocity splits into $\mathbf{v}^{\mathrm{cmd}}$ and $\mathbf{v}^{\mathrm{res}}=\mathbf{v}^{\mathrm{aug}}$ (synthesis used $\mathbf{v}^{\mathrm{cmd}}=\mathbf{0}$), and its tracking target is $\mathbf{v}^{\mathrm{tar}}=\mathbf{v}^{\mathrm{cmd}}+\mathbf{v}^{\mathrm{res}}$ with the command freshly sampled at training time.

Observations are organized into four groups.
The deployable observation collects the standard proprioception and the commands,
$\mathbf{o}_t=\big[\boldsymbol{\omega}_b,\allowbreak \mathbf{g},\allowbreak \mathbf{q}_t,\allowbreak \dot{\mathbf{q}}_t,\allowbreak \mathbf{a}_{t-1},\allowbreak \mathbf{v}^{\mathrm{cmd}},\allowbreak h^{\mathrm{cmd}},\allowbreak \mathbf{q}^{\mathrm{ref}},\allowbreak \log K_{\mathrm{L}}^{\mathrm{rob}},\allowbreak \log K_{\mathrm{L}}^{\mathrm{rob},\theta},\allowbreak \log K_{\mathrm{R}}^{\mathrm{rob}},\allowbreak \log K_{\mathrm{R}}^{\mathrm{rob},\theta},\allowbreak \log K_{\mathrm{T}}^{\mathrm{rob}}\big]\in\mathbb{R}^{95}$,
with the base angular velocity $\boldsymbol{\omega}_b$, the projected gravity $\mathbf{g}$, the joint angles $\mathbf{q}_t$ and velocities $\dot{\mathbf{q}}_t$, and the previous action $\mathbf{a}_{t-1}$.
The actor consumes the 10-frame history $\mathbf{o}_{t-9:t}$.
The privileged observation
$\mathbf{o}^{\mathrm{priv}}_t=\big[R_b,\allowbreak \dot{\mathbf{p}}_b,\allowbreak \{\mathbf{f}_j^{\mathrm{ext}},\boldsymbol{\tau}_j^{\mathrm{ext}}\}_{j=1}^{N},\allowbreak \mathbf{v}^{\mathrm{res}},\allowbreak h^{\mathrm{res}},\allowbreak \mathbf{q}^{\mathrm{res}}\big]\in\mathbb{R}^{88}$
includes the ground-truth base orientation $R_b$ (as a unit quaternion) and linear velocity $\dot{\mathbf{p}}_b$, the external wrenches applied at the upper-body links, and the three residuals.
\revtext{During training, Gaussian noise is added to the proprioception in $\mathbf{o}_t$, while $\mathbf{o}^{\mathrm{priv}}_t$ is noise-free.}
The base angular velocity, base linear velocity, velocity command, and external wrenches are all expressed in the base frame.
The encoder consumes the 5-frame history $\mathbf{o}^{\mathrm{priv}}_{t-4:t}$.
The target velocity also decides the locomotion mode.
\revtext{A frame is labeled stance when all three components of $\mathbf{v}^{\mathrm{tar}}$ are at most $0.1$ in magnitude, and walk otherwise.}
Determined jointly by the command and the privileged residual, this label is provided as the third, critic-only group $\mathbf{o}^{\mathrm{crit}}_t$.
The asymmetric critic observes $\mathbf{o}_{t-9:t}$, $\mathbf{o}^{\mathrm{priv}}_{t-4:t}$, and $\mathbf{o}^{\mathrm{crit}}_{t-4:t}$.
The fourth group is the 64-frame (1.28~s) history $\mathbf{o}_{t-63:t}$, consumed only by the estimator below.

\textbf{Teacher--student training.}
We adopt an RMA-style two-stage teacher--student scheme~\cite{kumar2021rma}.
In the first stage, a privileged encoder $\mathcal{E}^{\mathrm{priv}}$ compresses $\mathbf{o}^{\mathrm{priv}}_{t-4:t}$ into a latent $Z_t\in\mathbb{R}^{64}$, and the actor $\pi$, conditioned on $\mathbf{o}_{t-9:t}$ and $Z_t$, is trained with PPO~\cite{schulman2017proximal} together with the encoder and an asymmetric critic $V$.
In parallel, an estimator $\mathcal{E}^{\mathrm{est}}$ is regressed to reproduce $Z_t$ from the 64-frame history $\mathbf{o}_{t-63:t}$.
In the second stage, the encoder and actor are frozen, rollouts are collected with the actor conditioned on the estimate $\hat{Z}_t$, and the regression continues.
The deployed controller is the estimator--actor pair, driven purely by proprioception and commands.
\revtext{On the robot, $\boldsymbol{\omega}_b$ and $\mathbf{g}$ come from the onboard IMU.}
The actor runs at 50~Hz and outputs $\mathbf{a}_t\in\mathbb{R}^{23}$, which is scaled and added to the default posture to form the target angles of all 23 joints, tracked by joint-level PD controllers.

\textbf{Rewards.}
The reward sums three groups.
Tracking terms realize the augmented motion: Gaussian kernels on the base velocity against $\mathbf{v}^{\mathrm{tar}}$, the base height against $h^{\mathrm{aug}}$, and each of the eleven upper-body joints against $\mathbf{q}^{\mathrm{aug}}$.
Gait terms are standard bipedal shaping~\cite{zhang2025falcon} (stride period, swing height, contact schedule, foot placement), routed by the stance/walk label.
In stance, the walk-only terms are disabled, the velocity target is zeroed, and the robot is rewarded for keeping both feet planted.
Auxiliary terms regularize the motion: base orientation, energy, action smoothness, penalties on approaching the joint limits, and a survival bonus.

\textbf{Training details.}
Training runs in Isaac Lab~\cite{mittal2025isaac} on four GPUs simulating 8{,}192 environments each (32{,}768 in total) at about 32~GB of memory per GPU; the two stages take 20{,}000 and 10{,}000 iterations.
\revtext{Because the augmentation is computed offline, it does not increase the time of each training iteration.}

\revtext{In the first stage, a curriculum over the summed peak force or couple magnitude of each episode starts from the weakest wrenches and unlocks stronger ones as the tracking rewards grow, and the second stage uses all of them.}
\revtext{We also apply symmetry augmentation~\cite{mittal2024symmetry} during the PPO updates and sample each episode with a weight decreasing with the event count of its rarest contact link, mixed with a uniform floor.}

\section{Experiments}

\revtext{The simulation tests in Secs.~\ref{ssec:effect} to~\ref{ssec:ablation} are conducted in MuJoCo and, unless otherwise stated, share the following setup.}
\revtext{The velocity command $\mathbf{v}^{\mathrm{cmd}}$ is zero.}
\revtext{Small random perturbations are added to the initial joint angles and velocities of each run.}
\revtext{The target of a contact link is the steady state of its virtual admittance (Sec.~\ref{ssec:aug}).}
\revtext{Its displacement along the force is $\|\mathbf{f}_j^{\mathrm{ext}}\|/K_j^{\mathrm{rob}}$, and its rotation is $\|\boldsymbol{\tau}_j^{\mathrm{vir}}\|/K_j^{\mathrm{rob},\theta}$ along $\boldsymbol{\tau}_j^{\mathrm{vir}}$.}
\revtext{The IK label is the pose of the link in the augmented motion solved by IK.}
\revtext{During the tests, the CoM displacement and the attitude change of the link are measured in the world frame relative to its pose before the wrench is applied.}
\revtext{Both are projected onto the corresponding directions of the target and averaged over the last 1~s of the hold phase.}

\subsection{Effectiveness of Linear and Angular Compliance}
\label{ssec:effect}
\begin{figure}[t]
\centering
\includegraphics[width=0.88\columnwidth]{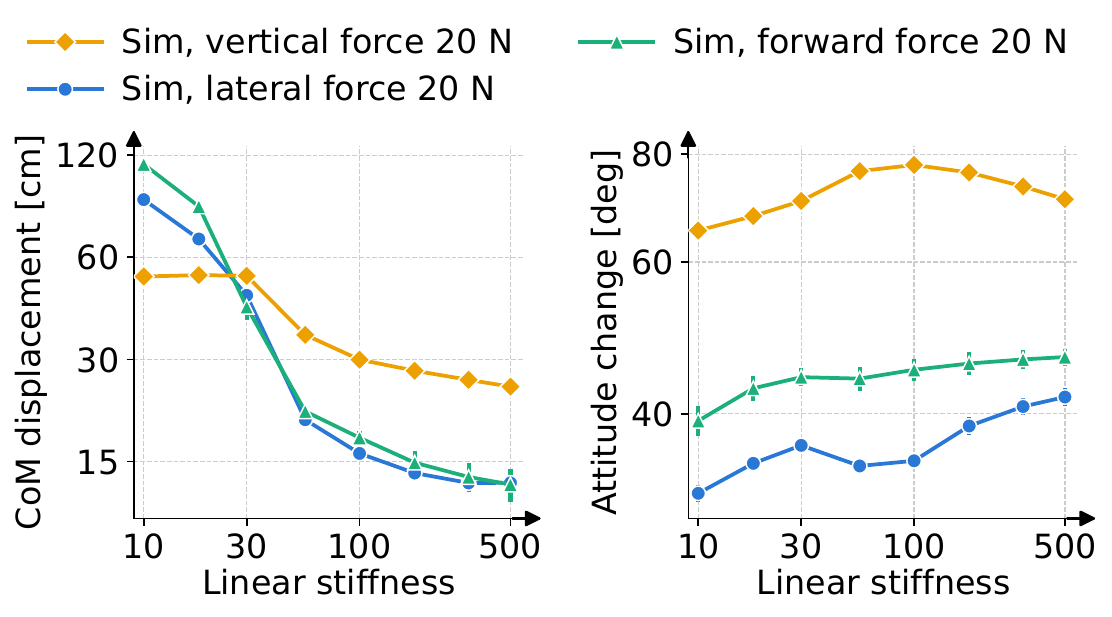}
\caption{\revtext{\textbf{Effect of the linear stiffness command.}} \revtext{CoM displacement (left) and attitude change (right) of the hand link versus the commanded linear stiffness of its arm in N/m; axes in log scale; error bars: s.d.\ ($\le 3.6$~cm, $\le 1.6^\circ$).}}
\label{fig:klin}
\end{figure}

We first test whether the policy modulates the pose of the contact link continuously according to the stiffness command.
Fixing the other four dimensions of the command, we vary the left-arm linear stiffness over its full command range $[10,500]$~N/m.
Three cases are tested.
Each case applies a force with a 20~N peak at the hand along a different direction, forward, lateral, or vertically downward, and starts from a different initial posture.
\revtext{We plot the mean response of the hand link over three runs.}
As shown in Fig.~\ref{fig:klin}, the CoM displacement of the link decreases monotonically as the linear stiffness grows, while the attitude change remains nearly constant.
Notably, the displacement is modulated over a narrower range in the vertical direction.
\revtext{At the soft end, remaining upright limits the drop of the link's CoM to about 53~cm.}
\revtext{At the stiff end, the hand rotates by about $70^\circ$ under the arm's angular stiffness of 10~N\,m/rad, which also lowers its CoM.}

\begin{figure}[t]
\centering
\includegraphics[width=0.88\columnwidth]{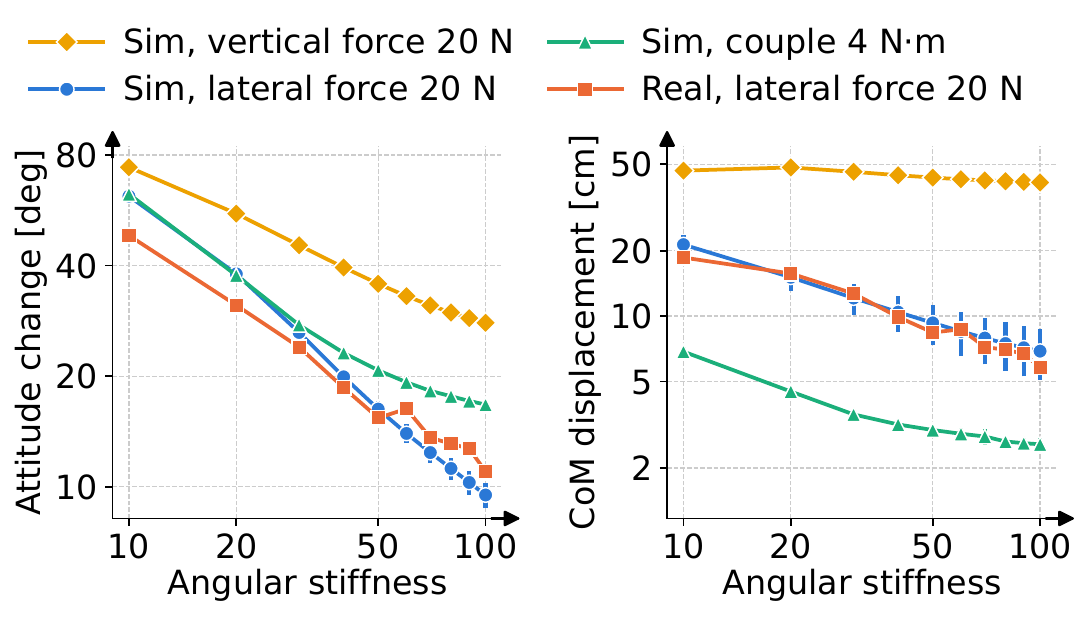}
\caption{\revtext{\textbf{Effect of the angular stiffness command.}} \revtext{Attitude change (left) and CoM displacement (right) of the hand link versus the commanded angular stiffness of its arm in N\,m/rad; axes in log scale; error bars: s.d.\ \mbox{($\le 2.3$~cm, $\le 1.2^\circ$)}.}}
\label{fig:kang}
\end{figure}
\begin{figure}[t]
\centering
\includegraphics[width=\columnwidth]{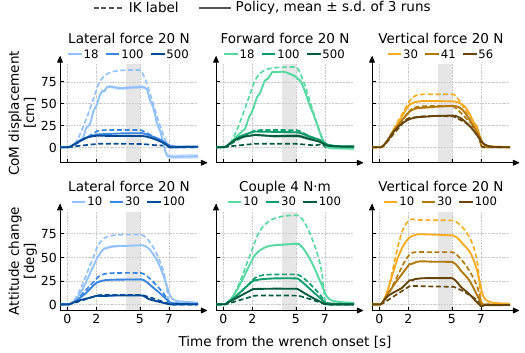}
\caption{\revtext{\textbf{IK labels and achieved trajectories.}} \revtext{Tests of Figs.~\ref{fig:klin} (top) and~\ref{fig:kang} (bottom) at three stiffness commands each (light to dark; N/m top, N\,m/rad bottom).} \revtext{Policy s.d.\ $\le 3.3$~cm and $\le 1.5^\circ$; gray: last 1~s of the hold phase.}}
\label{fig:traj}
\end{figure}

We next vary the arm angular stiffness over its full command range $[10,100]$~N\,m/rad.
Three cases are tested: a 20~N vertical downward force at the left hand, a 4~N\,m couple applied to the left hand about the wrist-roll joint's axis, and a 20~N horizontal rightward pull at the right palm.
\revtext{For the couple, which has no force direction, the CoM displacement is its magnitude.}
\revtext{As shown in Fig.~\ref{fig:kang}, the attitude change of the contact link decreases monotonically as the angular stiffness command grows, while the CoM displacement changes much less.}
\revtext{In the lateral-pull test, the displacement decreases by about 14~cm as the angular stiffness rises.}
\revtext{This is because the IK label relies increasingly on moving the base along the pull to realize the translation.}
\revtext{However, the resulting velocity target $\mathbf{v}^{\mathrm{tar}}$ is within the stance threshold (Sec.~\ref{ssec:rl}).}
\revtext{The policy, trained to track a zero velocity in stance, keeps its base stationary.}
We further replicate the lateral-pull test on the real robot, where a person pulls the right palm through a force gauge.
\revtext{The measurements, taken at the peak pull in one trial at each angular stiffness, are plotted in orange in Fig.~\ref{fig:kang}.}
\revtext{They match the simulation (blue) in both trend and range, indicating that the policy transfers to the real robot.}

\revtext{Fig.~\ref{fig:traj} shows the IK labels and the achieved trajectories of the six simulation tests.}
\revtext{Each test is repeated with a 3-s hold phase at three stiffness commands.} \revtext{These are the softest and stiffest commands whose IK labels pass the feasibility check at the full wrench, and one about midway between them on a log scale.}
\revtext{During the ramp-up, hold, and release phases of the wrench, the responses follow the trends of their IK labels, and after the wrench has returned to zero, they settle close to the initial pose.}
\revtext{In the last 1~s of the hold phase, the responses at the softest and middle stiffness commands match or are smaller than their labels.}
\revtext{At the stiffest commands, the responses match the labels under the vertical force of Fig.~\ref{fig:klin} and the lateral force of Fig.~\ref{fig:kang}, but are larger in the other four tests.}

\revtext{We further quantify the coupling on force episodes synthesized with a new random seed.}
\revtext{For each arm link under force, the linear or the angular stiffness of the link's arm is resampled, and a new IK label is synthesized under the same force.}
\revtext{At the 1,227 links where the new IK labels pass the feasibility check for both the linear and the angular stiffness, the policy is run twice under the same force, once with the original commands and once with the resampled command.}
\revtext{Both runs use the same velocity command, which is zero in 40\% of the episodes and a walking command in the others.}
\revtext{While the robot walks, the force rotates with the base yaw, and the IK label is moved by the integral of the velocity target $\mathbf{v}^{\mathrm{tar}}$.}
\revtext{When the resampled command differs from the original by a factor of two or more, the displacement of the link differs between the two runs by a median of 17.7~cm for a resampled linear stiffness (680 links), versus 2.9~cm for a resampled angular stiffness (555 links).}
\revtext{Its attitude change differs by a median of $20.5^\circ$ for a resampled angular stiffness, versus $3.8^\circ$ for a resampled linear stiffness.}
\revtext{Overall, the two commands can be adjusted separately, although there is a minor coupling between their effects on the response.}
\revtext{The pose changes caused by this coupling mainly come from the execution errors of the policy, such as the tracking error of the base velocity.}

\begin{table}[t]
\centering
\caption{\revtext{Response errors for one and two contacts.}}
\label{tab:multicontact}
\setlength{\tabcolsep}{3.5pt}\renewcommand{\arraystretch}{1.25}\scriptsize
\begin{tabular}{lcccccc}
\hline
& \multicolumn{2}{c}{Single} & \multicolumn{2}{c}{Two} & \multicolumn{2}{c}{Force and couple}\\[-2pt]
& \multicolumn{2}{c}{force} & \multicolumn{2}{c}{forces} & \multicolumn{2}{c}{on the same hand}\\
\cmidrule(lr){2-3}\cmidrule(lr){4-5}\cmidrule(lr){6-7}
& Standing & Walking & Standing & Walking & Standing & Walking\\
\hline
Position [cm] & 13.5 & 12.9 & 9.6 & 9.1 & 16.1 & 13.6\\
Orientation [deg] & 11.2 & 12.0 & 9.3 & 9.8 & 13.3 & 14.9\\
\hline
\end{tabular}
\end{table}

\revtext{We also evaluate the policy on episodes that contain two forces or a force and a couple on the same hand.}
\revtext{The latter combination does not occur in training.}
\revtext{For each episode type, 800 episodes are synthesized with a new random seed.}
\revtext{The two forces are on different arms, and the force and the couple on the same hand have the same time profile.}
\revtext{Episodes that contain a single force on an arm are evaluated in the same way.}
\revtext{Each episode is evaluated twice, under the zero velocity command and under a walking command.}
\revtext{Table~\ref{tab:multicontact} reports the RMSE from the IK label to the policy response over all loaded links.}
\revtext{Under two forces, the errors are smaller than those under a single force.}
\revtext{This is because the feasibility check makes each of the two forces about half as large as a single force on average, so the IK labels are easier for the policy to realize.}
\revtext{Under a force and a couple on the same hand, the errors are larger than under a single force.}

\subsection{Comparison with Baseline Methods}

\begin{figure}[t]
\centering
\includegraphics[width=\columnwidth]{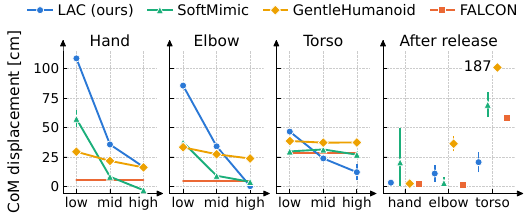}
\caption{\revtext{\textbf{Comparison with baseline methods.}} \revtext{The left three panels show the mean $\pm$ s.d.\ of 3 runs per setting.} \revtext{The right panel shows the mean $\pm$ s.d.\ of the absolute values over all settings, with the error bars clipped at zero.}}
\label{fig:quant}
\end{figure}
We compare LAC with the open-source baselines SoftMimic~\cite{margolis2025softmimic}, GentleHumanoid~\cite{lu2025gentlehumanoid}, and FALCON~\cite{zhang2025falcon}.
Three scenarios are tested: a 30~N forward pull at the left hand, a 30~N rightward pull at the right elbow, and a 20~N backward push on the torso.
\revtext{Each force ramps up over 1~s, holds for 1~s, and decreases to zero over 2~s.}
\revtext{We set the command of each method to the minimum, an intermediate value, and the maximum of its range.}
\revtext{For LAC, this command is the linear stiffness of the contact link's group, set to 10, 70, and 500~N/m.}
\revtext{For SoftMimic, it is the linear stiffness, set to 40, 180, and 800~N/m.}
\revtext{For GentleHumanoid, it is the force threshold, set to 5, 10, and 15~N, whose higher values correspond to stiffer behavior.}
\revtext{FALCON serves as a fixed baseline.}
\revtext{We therefore compare the adjustable response ranges of the methods.}
\revtext{Fig.~\ref{fig:quant} reports the CoM displacement of the contact link, averaged over the last 0.5~s of the hold phase, and the displacement that remains after the force has returned to zero.}

\revtext{LAC is the only method whose displacement decreases monotonically with its command in all three scenarios.}
\revtext{SoftMimic and GentleHumanoid adjust the displacement over narrower ranges, and their commands have little effect under the torso push, which is applied outside their wrench application sites (Table~\ref{tab:capability}).}
\revtext{FALCON provides a single response in each scenario.}
\revtext{After the force has returned to zero, the mean remaining displacement of LAC is 3, 11, and 21~cm in the three scenarios, whereas each baseline leaves a much larger displacement in at least one scenario.}

\subsection{\texorpdfstring{\revtext{Ablation of the Angular Target Construction}}{Ablation of the Angular Target Construction}}
\label{ssec:ablation}

\begin{figure}[t]
\centering
\includegraphics[width=\columnwidth]{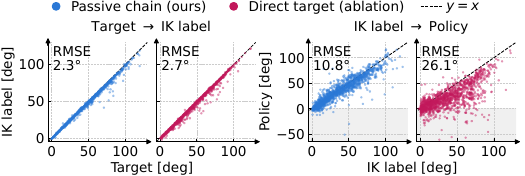}
\caption{\revtext{\textbf{Angular target ablation on augmented episodes unseen in training.}}
\revtext{Each dot is one arm contact, with rotations signed along the respective target direction.}
\revtext{The shaded areas mark rotations opposite to the target.}
\revtext{RMSE is taken about $y=x$.}}
\label{fig:ablation_scatter}
\end{figure}
\begin{figure}[t]
\centering
\includegraphics[width=0.94\columnwidth]{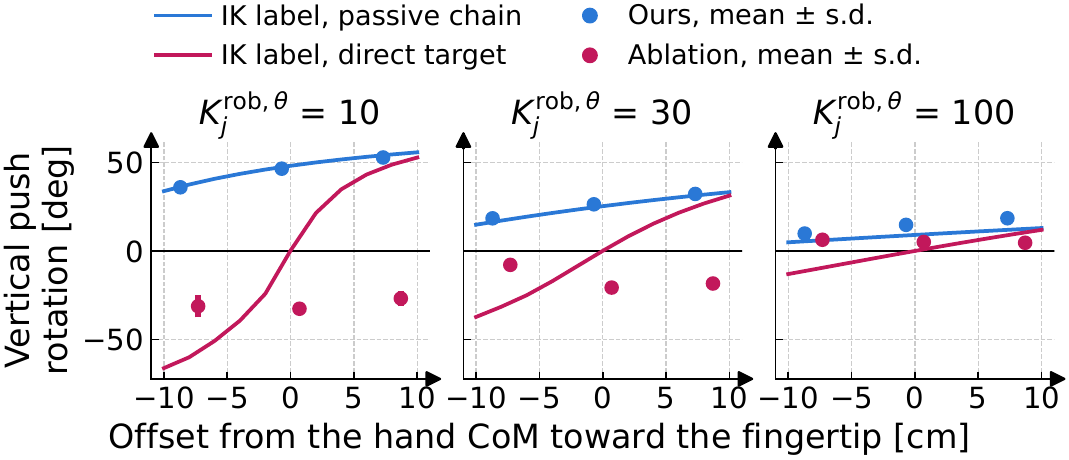}
\caption{\revtext{\textbf{Angular target ablation across application points.}}
\revtext{A 20~N force pushes the left hand down.}
\revtext{Positive rotation follows the passive arm.}
\revtext{Lines: IK labels; markers: mean of 3 runs, s.d.\ $\le 1.1^\circ$ (ours), $\le 6.3^\circ$ (ablation).}}
\label{fig:ablation_sweep}
\end{figure}
\revtext{To isolate the passive-chain construction, we synthesize an ablation dataset in which the force events use a direct target.}
\revtext{In the direct target, the virtual torque is replaced by $\boldsymbol{\tau}_j^{\mathrm{dir}}=\gamma\,\mathbf{r}_j\times\mathbf{f}_j^{\mathrm{ext}}$, the torque of the applied force about the link CoM scaled by $\gamma=11.6$.}
\revtext{Here, $\mathbf{r}_j$ points from the CoM to the application point.}
\revtext{Couple events, the IK configuration, and the training settings are unchanged, and an ablation policy is trained on this dataset.}

\revtext{We first evaluate both policies on about 1{,}000 force episodes synthesized with a new random seed.}
\revtext{As shown in Fig.~\ref{fig:ablation_scatter}, the IK realizes both kinds of targets with similar accuracy.}
\revtext{In contrast, the RMSE from the IK label to the policy response is much smaller for LAC than for the ablation.}

\revtext{We then apply a 20~N force at offsets of $-8$, $0$, and $+8$~cm from the hand CoM (positive toward the fingertip), pushing the left hand down (Fig.~\ref{fig:ablation_sweep}).}
\revtext{Across these points, the IK labels of the passive chain remain positive, whereas those of the direct target change sign when the application point crosses the CoM.}
\revtext{Despite this sign change, the ablation policy rotates the hand in the same direction at all three points.}
\revtext{Overall, most direct targets are kinematically reachable, but the policy realizes them with substantially larger errors than the passive-chain targets.}

\subsection{Performance in Real-World Tasks}
\begin{figure}[t]
\centering
\includegraphics[width=0.97\columnwidth]{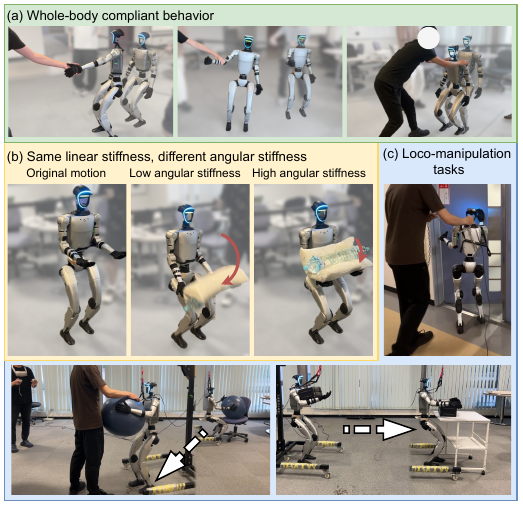}
\caption{\revtext{\textbf{Real-world performance of LAC.}}
\revtext{\textbf{(a)} The robot responds to external wrenches applied at the hand, elbow, and torso with whole-body compliant behavior.}
\revtext{\textbf{(b)} Under the same linear stiffness, the angular stiffness command alone reshapes the response.}
\revtext{\textbf{(c)} Loco-manipulation tasks under VR teleoperation.}}
\label{fig:bigshow}
\end{figure}
We further evaluate LAC extensively in the real world.
In the trials of Fig.~\ref{fig:bigshow}(a), the contact link is commanded with low linear and angular stiffness.
In the first snapshot, pulling the left hand forward makes the robot extend the arm and step forward.
In the second, pulling the right elbow sideways makes the robot raise the elbow and move to the right.
In the third, pressing down on the torso makes the robot bend its knees and lower its height.

The modulation of the angular stiffness is important for tasks whose success depends on the attitude of the contact links, as shown in Fig.~\ref{fig:bigshow}(b).
With a 4~kg object placed on the arms, a low angular stiffness (10) fails to keep the arms level and the object falls off, whereas a high angular stiffness (100) carries it steadily.

\revtext{Finally, in the loco-manipulation tasks of Fig.~\ref{fig:bigshow}(c), a human operator provides both the motion and the stiffness commands through VR teleoperation.}
Low linear and angular stiffness on both arms facilitates the transport of soft objects such as a yoga ball.
\revtext{No fine adjustment of the motion commands is then required to keep contact with the ball without deforming it much.}
Conversely, high stiffness on both arms enables force-demanding tasks, such as carrying a heavy box and placing it on a table, or pushing open a spring-loaded door.

\section{Conclusion}
We presented LAC, a humanoid whole-body controller that realizes commanded linear and angular compliance through large-scale compliant motion augmentation and teacher--student reinforcement learning.
Looking ahead, two aspects leave room for improvement.
\revtext{First, the policy perceives the external wrench only through proprioception, and measuring the wrenches on the hands with force/torque sensors could enable fine-grained force control.}
\revtext{Second, designing more types of wrench events, such as a force and a couple applied together, could improve the compliant response of the policy in a wider range of interactions.}

\bibliographystyle{IEEEtran}
\bibliography{references}

\end{document}